\documentclass[10pt,twocolumn,letterpaper]{article}

\usepackage{cvpr}
\usepackage{times}
\usepackage{epsfig}
\usepackage{graphicx}
\usepackage{amsmath}
\usepackage{amssymb}
\usepackage{multirow}
\usepackage{color}
\usepackage{lipsum}
\usepackage{bbding}

\usepackage[breaklinks=true,bookmarks=false]{hyperref}

\cvprfinalcopy 

\def\cvprPaperID{****} 

\begin{document}
\title{Pyramid Feature Attention Network for Saliency detection}

\author{Ting Zhao, Xiangqian Wu\\
School of Computer Science and Technology, Harbin Institute of Technology \\
{\tt\small 17S003073@stu.hit.edu.cn, xqwu@hit.edu.cn\Envelope}
}

\maketitle

\begin{abstract}
Saliency detection is one of the basic challenges in computer vision. How to extract effective features is a critical point for saliency detection. Recent methods mainly adopt integrating multi-scale convolutional features indiscriminately. However, not all features are useful for saliency detection and some even cause interferences. To solve this problem, we propose Pyramid Feature Attention network to focus on effective high-level context features and low-level spatial structural features. First, we design Context-aware Pyramid Feature Extraction (CPFE) module for multi-scale high-level feature maps to capture rich context features. Second, we adopt channel-wise attention (CA) after CPFE feature maps and spatial attention (SA) after low-level feature maps, then fuse outputs of CA $\&$ SA together. Finally, we propose an edge preservation loss to guide network to learn more detailed information in boundary localization. Extensive evaluations on five benchmark datasets demonstrate that the proposed method outperforms the state-of-the-art approaches under different evaluation metrics.
\end{abstract}

\section{Introduction}

Saliency detection aims to locate the important parts of natural images which attract our attention. As the pre-processing of computer vision applications, e.g. object detection\cite{retarget1, retarget2}, visual tracking\cite{track1,track2}, image retrieval\cite{retrieval1,retrieval2} and semantic segmentation\cite{segment}, saliency detection attracts many researchers. Currently, the most effective saliency detection methods are based on the fully convolutional network (FCN). FCN stacks multiple convolution and pooling layers to gradually increase the receptive field and generate the high-level semantic information, which plays a crucial role in saliency detection. However, the pooling layers reduce the size of the feature maps and deteriorate the boundaries of the salient objects.

\begin{figure}[h]
\centering
\includegraphics[width=1.0\linewidth]{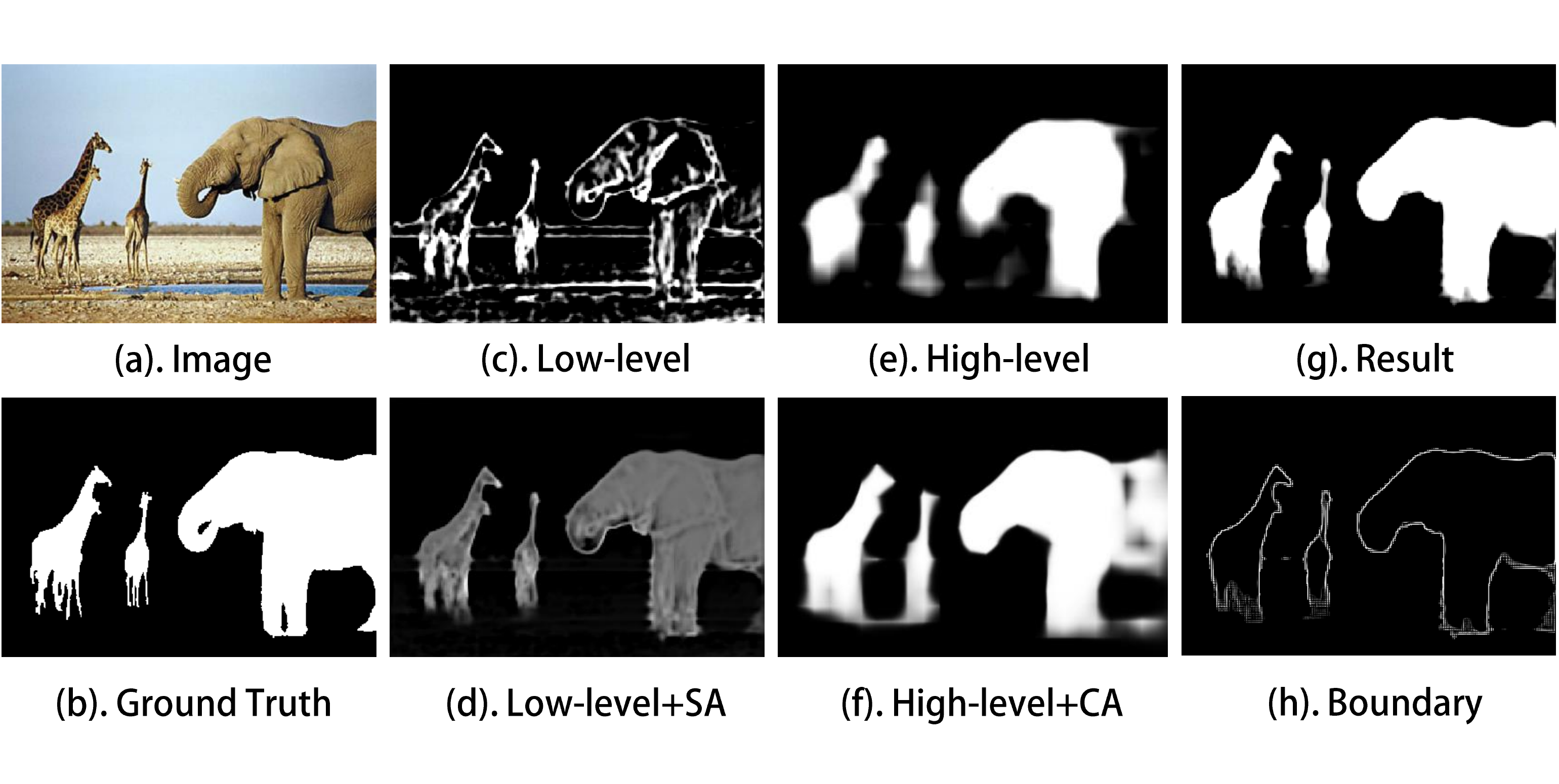}
\caption{An example of applying Pyramid Feature Attention network.(a) and (b) represent the input image and corresponding Ground Truth. (c) and (d) mean low-level features without or with spacial attention.  (e) and (f) are high-level features without or with channel-wise attention. (g) and (h) represent the results from our method and the boundary map of (g) generated by Laplace operator.}
\label{fig:Figure_1}
\end{figure}

To deal with this problem, some works introduce hand-craft features to preserve the boundaries of salient objects\cite{eld,region_pix}. \cite{eld} extracts the hand-craft features to compute the salient values of super-pixels. \cite{region_pix} partitions the image into regions by hand-craft features. When generating saliency maps, the hand-craft features and the CNN high-level features are complementary but extracted separately in these methods. However,  it is difficult to effectively fuse the complementary features extracted separately. Furthermore, hand-craft features extraction is a time-consuming procedure.

Besides hand-craft features, some works discover that the features from different layers of the network are also complementary and integrate the multi-scale features for saliency detection \cite{dss,amulet,dsrn}. More specifically, the features at deep layers typically contain global context-aware information, which are suitable to locate the salient regions correctly. The features at shallow layers contain the spatial structural details, which are suitable to locate boundaries. These methods fused different scale features without considering their different contribution for saliency, it is not optimal for saliency detection. To overcome these problems, attention model \cite{pagrn} and gate function \cite{bdmpm} are introduced to the saliency detection networks. However, the methods ignore the different characteristics of the high-level and low-level features, which may affect the extraction of effective features.

In this paper, we propose a novel salient object detection method named Pyramid Feature Attention (PFA) network. In consideration of the different characteristics of different level features (Fig.\ref{fig:Figure_1} (c,e)), the saliency maps from low-level features contain many noises, while the saliency maps from high-level features only get an approximate area. Therefore, for high-level features, inspired by SIFT\cite{sift} feature extraction algorithm, we design a context-aware pyramid feature extraction(CPFE) module to get multi-scale multi-receptive-field high-level features, and then we use channel-wise attention(CA) to select appropriate scale and receptive-field for generating saliency regions. In training process, CA assigns large weights to the channels which play important role for saliency detection(Fig.\ref{fig:Figure_1} (f)). In order to refine the boundaries of saliency regions, we fuse low-level features with edge information. But not all edge information is effective for refining saliency maps, we expect to focus on the boundaries between salient objects and background. So we use spatial attention to better focus on the effective low-level features, and obtain clear saliency boundaries(Fig.\ref{fig:Figure_1} (d)). After the processing of different attention mechanisms, the high-level features and low-level features are complementary-aware and suitable to generate saliency map.  In addition, different from previous saliency detection approaches, we propose an edge preservation loss to guide network to learn more detailed information in boundary localization. With the above considerations, the proposed method PFA network can produce good saliency maps.

In short, our contributions are summarized as follows: 

1. We propose a Pyramid Feature Attention (PFA) network for image saliency detection. For high-level feature, we adopt a context-aware pyramid feature extraction module and a channel-wise attention module to capture rich context information. For low-level feature, we adopt spatial attention module to filter out some background details. 

2. We design a novel edge preservation loss to guide network to learn more detailed information in boundary localization. 

3. The proposed model achieves the state-of-the-art on several challenging datasets. The experiments prove the effectiveness and superiority of the proposed method.

\section{Related Works}

\subsection{Salient Object Detection}
In the past decade, there are a number of approaches for saliency detection. Early approaches\cite{craft1,craft2,craft3,craft4} estimate the salient value based on hand-crafted features. Those approaches detect salient objects with humanlike intuitive feelings and heuristic priors, such as color contrast\cite{craft1}, boundary background\cite{craft2,craft3} and center prior\cite{craft4}. These direct techniques are known to be friendly to keep fine image structure. Nevertheless, the hand-craft features and priors can hardly capture high-level and global semantic knowledge about the objects.

In recent years, many efforts about various network architectures have been made in saliency detection.  Some experiments\cite{dss,eld,dsrn} show that high-level features in deep layers encode the semantic information for getting an abstract description of objects, while low-level features in shallow layers keep spatial details for reconstructing the object boundaries (Fig.\ref{fig:Figure_1} (c,e)). Accordingly, some works bring multi-level features into saliency detection. Hou et al. \cite{dss} propose a saliency method by introducing short connections to the skip-layer structures within the HED architecture. Wang et al. \cite{rfcn} propose a saliency detection method based on recurrent fully convolutional networks (RFCNs). Luo et al. \cite{nldf} combine the local and global information through a multi-resolution grid structure. Zhang et al. \cite{amulet} aggregate multi-level features by concatenating feature maps from both high levels and low levels directly. Zhang et al. \cite{bdmpm} propose a bi-directional message passing module, where messages can transmit mutually controlled by gate function. However, some features may cause interferences in saliency detection. How to get various features and select effective ones becomes an important problem in saliency detection.

\subsection{Attention Mechanisms}
Attention mechanisms have been successfully applied in various tasks such as machine translation \cite{translation}, object recognition \cite{detection1}, image captioning \cite{scacaption,caption2}, visual question answering \cite{qa1, qa2} and pose estimation \cite{pose}. Chu et al. \cite{pose} propose a network with multi-context attention mechanism into an end-to-end framework for human pose estimation. Chen et al. \cite{scacaption} propose a SCA-CNN network that incorporates spatial and channel-wise attention in CNN for image captioning. Zhang et al.\cite{pagrn} propose a progressive attention guided network which generates attentive features by channel-wise and spatial attention mechanisms sequentially for saliency detection.

\begin{figure*}[t]
\centering
\includegraphics[width=1.0\linewidth]{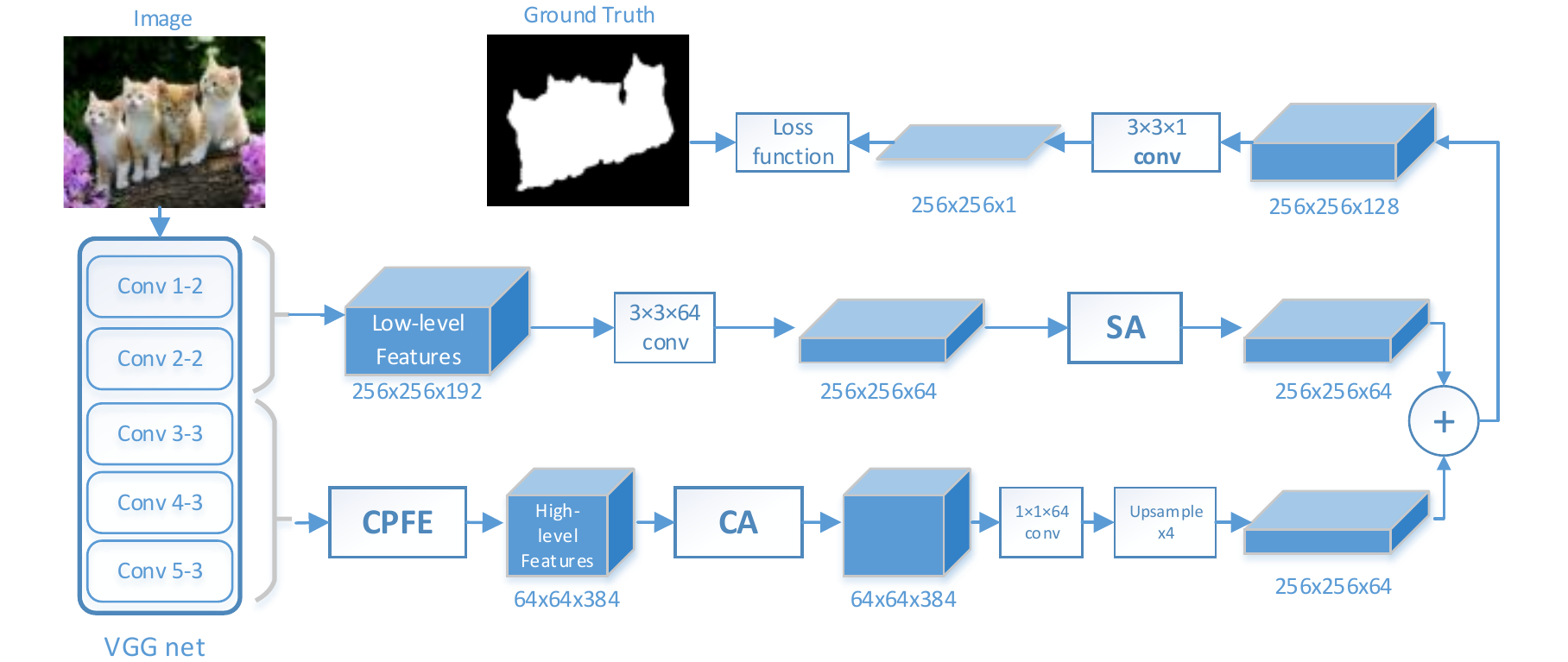}
\caption{The overall architecture of our method. CPFE means context-aware pyramid feature extraction. The high-level features are from vgg3-3, vgg4-3 and vgg5-3. The low-level features are from vgg1-2 and 2-2, which upsample to the size of vgg1-2.}
\label{fig:Figure_2}
\end{figure*}

Due to attention mechanisms have great ability to select features, it is a perfect fit for saliency detection. While integrating the convolutional features, most existing methods treat multi-level features without distinction. Some methods adopted certain valid strategies, such as gate function\cite{bdmpm} and progressive attention\cite{pagrn}, but those methods select features in a certain direction and ignore the differences between high-level and low-level features. Different from them, for high-level feature, we adopt context-aware pyramid feature extraction(CPFE) module and channel-wise attention module  to capture rich context information. In CPFE module, we adopt multi-scale atrous convolutions on the side of three high-level blocks of VGG net, then channel-wise attention mechanism assigns large weights to channels which show high response to salient objects. For low-level feature, there exists some background regions which distract the generation of saliency map. Spatial attention mechanism filters out some background details according to high-level features and focus more on the foreground regions, which helps to generate effective features for saliency prediction. 

\section{Pyramid Feature Attention Network}
In this paper, we propose a novel saliency detection method, which contains a context-aware pyramid feature extraction module and a channel-wise attention module to capture context-aware multi-scale multi-receptive-field high-level features, a spatial attention module for low-level feature maps to refine salient object details and an effective edge preservation loss to guide network to learn more detailed information in boundary localization. The overall architecture is illustrated in Fig.\ref{fig:Figure_2}.

\subsection{Context-aware pyramid feature extraction}
Visual context is quite important for saliency detection. Existing CNN models learn features of objects by stacking multiple convolutional and pooling layers. However, the salient objects have large variations in scale, shape and position. Previous methods usually directly use the bottom-to-up convolution and pooling layers, that may not be effectively to handle these complicated variations. Inspired by the feature extraction of SIFT\cite{sift}, we try to design a novel module to extract the features of scale, shape and location invariances. The scale-invariant feature transform (SIFT)  is a feature detection algorithm in computer vision to detect and describe local features in images. The algorithm proposed the Laplassian of Gaussian representation\cite{sift} which fused scale space representations and pyramid multi-resolution representations. The scale space representations which are processed by several different Gaussian kernel functions with same resolution; and the pyramid multi-resolution representations which are processed by down samples with different resolutions. Similar with Gaussian function in SIFT, we use atrous convolution \cite{deeplab} to get features with same scale but different receptive fields. Similar with pyramid multi-resolution representations in SIFT, we take conv3-3, conv4-3 and conv5-3 of VGG-16 \cite{vgg} to extract multi-scale features.

\begin{figure}[t]
\centering
\includegraphics[width=1.0\linewidth]{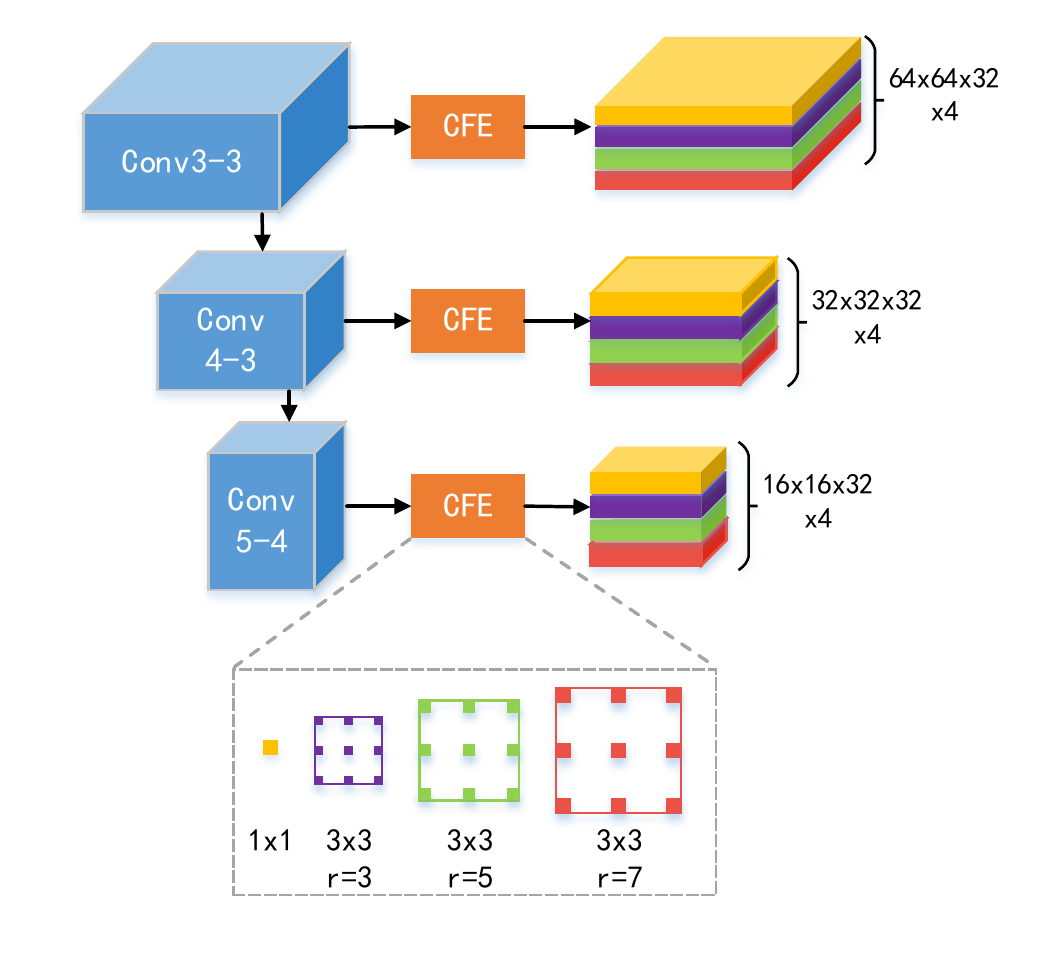}
\caption{ Detailed structure of context-aware pyramid feature extraction. A context-aware feature extraction module takes a feature from
a side output of  net as input and it contains three 3$\times$3 convolutional layers with different dilation rates and a 1$\times$1 convolutional layers, the output channel of each convolutional layer is 32.}
\label{fig:Figure_3_pyramid}
\end{figure}

Specifically, the context-aware pyramid feature extraction module is shown in Fig.\ref {fig:Figure_3_pyramid}. We take conv 3-3 , conv 4-3 and conv 5-3 in VGG-16 as the basic high-level features. To make the final extracted high-level features contain the features of scale and shape invariances, we adopt atrous convolution with different dilation rates, which are set to 3, 5 and 7 to capture multi-receptive-field context information. Then we combine the feature maps from different atrous convolutional layers and a 1$\times$1 dimension reduction feature by cross-channel concatenation. After this, we get three different scale features with context-aware information, we upsample the two smaller ones to the largest one. Finally, we combine them  by cross-channel concatenation as the output of the context-aware pyramid feature extraction module.

\subsection{Attention mechanism}
We exploit context-aware pyramid feature extraction to get multi-scale multi-receptive-field high-level features. Different features have different semantic values to generate saliency maps. But most existing methods integrate multi-scale features without distinction, which lead to information redundancy. More importantly, inaccurate information at some levels would lead to a performance degradation or even wrong prediction. It is significant to filter these features and fucus more on valuable features. In this subsection we will talk about the attention mechanisms in PFA network. According to the characteristics of different level features, we adopt channel-wise attention for high-level features and spacial attention for low-level features to select effective features. In addition, we don't use spacial attention for high-level features, because  high-level features contain high abstract semantics\cite{senet,pagrn}, there is no need to filter spacial information. While, we don't use channel-wise attention for low-level feature, because there are almost no semantic distinctions among different channels of low-level features.

\subsubsection{Channel-wise attention}
 Different channels of features in CNNs generate response to different semantics\cite{senet}. From Fig.\ref{fig:Figure_1}, the saliency map from high-level features is just a rough result, some essential regions may be weakened. We add channel-wise attention (CA) \cite{senet,scacaption} module after context-aware pyramid feature extraction to weighted multi-scale multi-receptive-field high-level features. The CA will assign larger weight to channels which show high response to salient objects . 

We unfold high-level features \textbf{\emph{f}}$^{h}$ $\in$ $\mathbb{R}^{W \times H \times C}$ as  \textbf{\emph{f}}$^{h}$= [ \textbf{\emph{f}}$_{1}^{h}$, \textbf{\emph{f}}$_{2}^{h}$,..., \textbf{\emph{f}}$_{C}^{h}$], where  \textbf{\emph{f}}$_{i}^{h}$ $\in$ $\mathbb{R}^{W \times H}$ is the \emph{i}-th slice of \textbf{\emph{f}}$^{h}$ and \emph{C} is the total channel number. First, we apply average pooling to each \textbf{\emph{f}}$_{i}^{h}$ to obtain a channel-wise feature vector \textbf{\emph{v}}$^{h}$ $\in$ $\mathbb{R}^{C}$. After that, two consecutive fully connected(FC) layer to fully capture channel-wise dependencies(see Fig.\ref {fig:Figure_attention} ). As \cite{senet}, to limit model complexity and aid generalisation, we encode the channel-wise feature vector by forming a bottleneck with two FC layers around the non-linearity. Then, through using sigmoid operation, we take the normalization processing measures to the encoded channel-wise feature vector mapped to [0,1].

\begin{equation}
CA=F(\textbf{\emph{v}}^{h},W)=\sigma_1 (fc_{2}(\delta(fc_{1}(\textbf{\emph{v}}^{h},W_{1})),W_{2}))
\end{equation}

Where $W$ refers to parameters in channel-wise attention block, $\sigma_1$ refers to sigmoid operation, $fc$ refers to FC layers, $\delta$ refers to the ReLU function. The final output $\widetilde f^h$ of the block is obtained by weighting the context-aware pyramid features with $CA$.
\begin{equation}
\widetilde f^h =CA\cdot f^h
\end{equation}

\subsubsection{Spacial attention}

Natural images usually contains a wealth of details of foreground and complex background. From Fig.\ref{fig:Figure_1}, the saliency map from low-level features contains a lot of details which easily brings bad results. In saliency detection, we want to obtain detailed boundaries between salient objects and background without other texture which can distract human attention. Therefore, instead of considering all spatial positions equally, we adopt spatial attention to focus more on the foreground regions, which helps to generate effective features for saliency prediction.

\begin{figure}[t]
\centering
\includegraphics[width=1.0\linewidth]{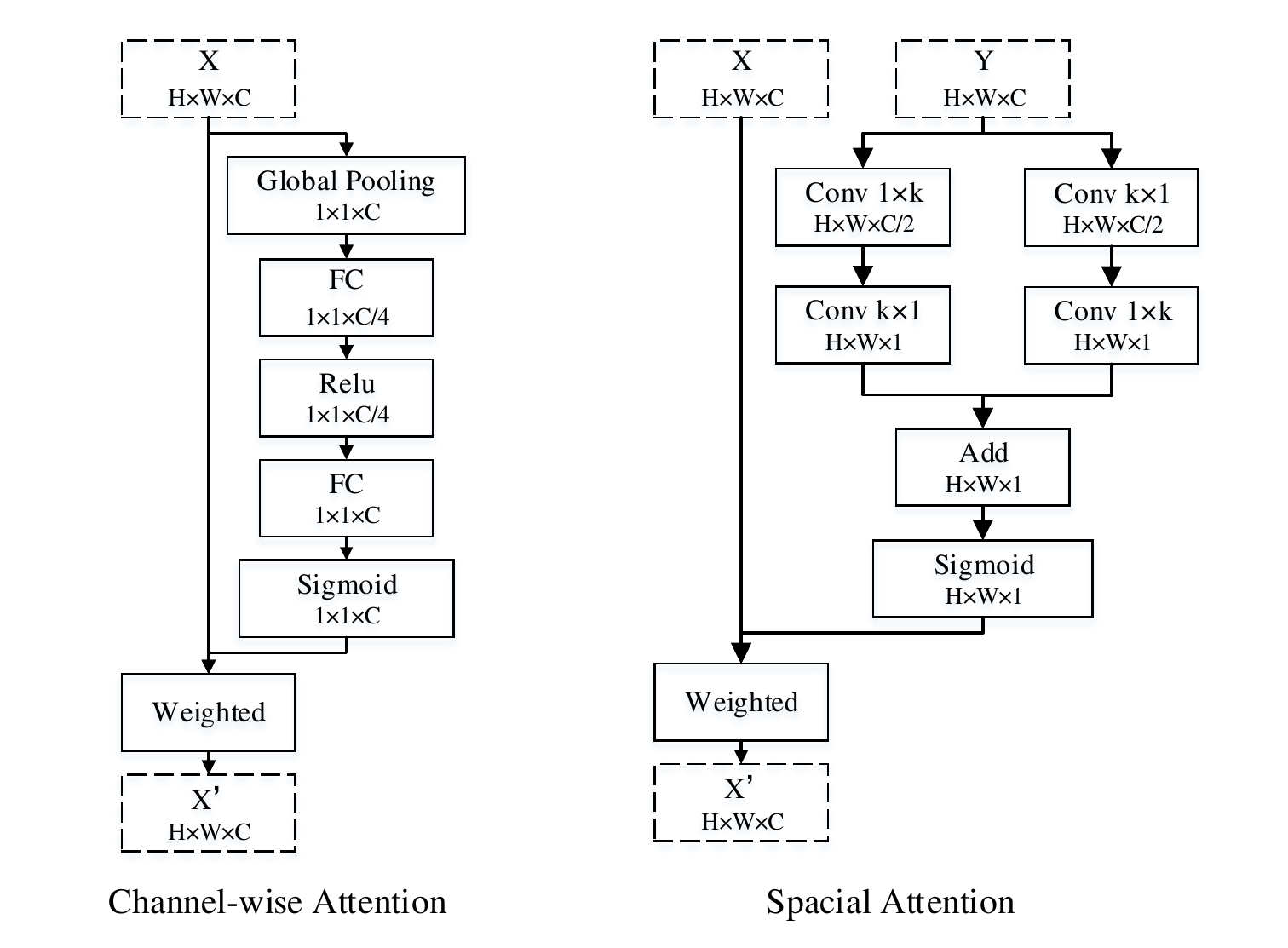}
\caption{ Channel-wise attention (left) and spacial attention (right). Where X and X' mean weighted feature and weighting feature respectively, Y means context-aware high-level feature after CA in this paper.}
\label{fig:Figure_attention}
\end{figure}

We represent low-level features as \textbf{\emph{f}}$^{l}$ $\in$ $\mathbb{R}^{W \times H \times C}$. The set of spatial locations is denoted by $\mathbb{R}=\{(x,y)|x=1,...,W; y=1,...,H\}$, where \emph{j} =(x,y) is the spatial coordinate of low-level features. For increasing receptive field and getting global information but not increasing parameters, similar to \cite{gcn}, we apply two convolution layers ,one's kernel is 1$\times$k and the other's is k$\times$1, for high-level feature to capture spacial concerns(see Fig.\ref {fig:Figure_attention} ). Then, using sigmoid operation, we take the normalization processing measures to the encoded spacial feature map mapped to [0,1].

\begin{equation}
C_1=conv_2(conv_1(\widetilde f^h,W_1^1),W_1^2))
\end{equation}

\begin{equation}
C_2=conv_1(conv_2(\widetilde f^h,W_2^1),W_2^2))
\end{equation}

\begin{equation}
SA=F(\widetilde f^h,W)=\sigma_2 (C_1 + C_2)
\end{equation}

Where $W$ refers to parameters in spacial attention block, $\sigma_2$ refers to sigmoid operation, $conv_1$ and $conv_2$ refers to 1$\times$k$\times$C and k$\times$1$\times$1 convlution layer respectively and we set k=9 in experiment. The final output $\widetilde f^l$ of the block is obtained by weighting $f^l$ with $SA$.
\begin{equation}
\widetilde f^l =SA\cdot f^l
\end{equation}

\subsection{Loss function}\label{sec:Loss function}
In machine learning and mathematical optimization, loss functions represent the price paid for inaccuracy of predictions in classification problems. In saliency object detection, we always use the cross-entropy loss between the final saliency map and the ground truth. The loss function is defined as:
\begin{equation}
\begin{split}
L_S=-\sum_{i=0}^{size(Y)}(\alpha_sY_{i}log(P_{i}) \\
+(1-\alpha_s)(1-Y_{i})log(1-P_{i}))
\end{split}
\end{equation}
where Y means the ground truth and P means the saliency map of network output, $\alpha_s$ means a balance parameter of positive and negative samples and we set $\alpha_s=0.528$ which calculated from groundtruth of the training set. However, the loss function just provides general guidance to generate saliency map. We use a simpler strategy to emphasize generation of the salient object boundaries details. First, we use Laplace Operator\cite{LO} to get boundaries of ground truth and saliency map of network output, and then we use the cross-entropy loss to supervise the generation of salient object boundaries.
\begin{equation}
\Delta f=\frac{{\partial}^{2}f}{\partial x^{2}}+\frac{{\partial}^{2}f}{\partial y^{2}}\label{Laplace_operator}
\end{equation}
 
\begin{equation}
\Delta \widetilde f=abs(tanh(conv(f,K_{laplace}))) \label{abstanh}
\end{equation}

\begin{equation}
\begin{split}
L_{B}=-\sum_{i=0}^{size(Y)}(\Delta Y_{i}log(\Delta P_{i}) \\
+(1- \Delta Y_{i})log(1- \Delta P_{i})) \label{cross-entropy_loss}
\end{split}
\end{equation}

The Laplace operator is a second order differential operator in the n-dimensional Euclidean space, defined as the divergence of the gradient ($\Delta f$). Because the second derivative can be used to detect edges, we use the Laplace operator to get salient object boundaries. The Laplace operator in two dimensions is given by Eq.\ref{Laplace_operator}, where x and y are the standard Cartesian coordinates of the xy-plane. In fact, since the Laplacian uses the gradient of images, it calls internally the convolution operation to perform its computation. Then we use absolute operation followed by tanh activatioin Eq.\ref{abstanh} map the value to [0,1]. Finally we use the cross-entropy loss to supervise the generation of salient object boundaries Eq.\ref{cross-entropy_loss}. The total loss function is their weighted sum:
\begin{equation}
L=\alpha L_{S}+(1-\alpha)L_{B}
\end{equation}

\begin{figure*}[t]
\centering
\includegraphics[width=1.0\linewidth]{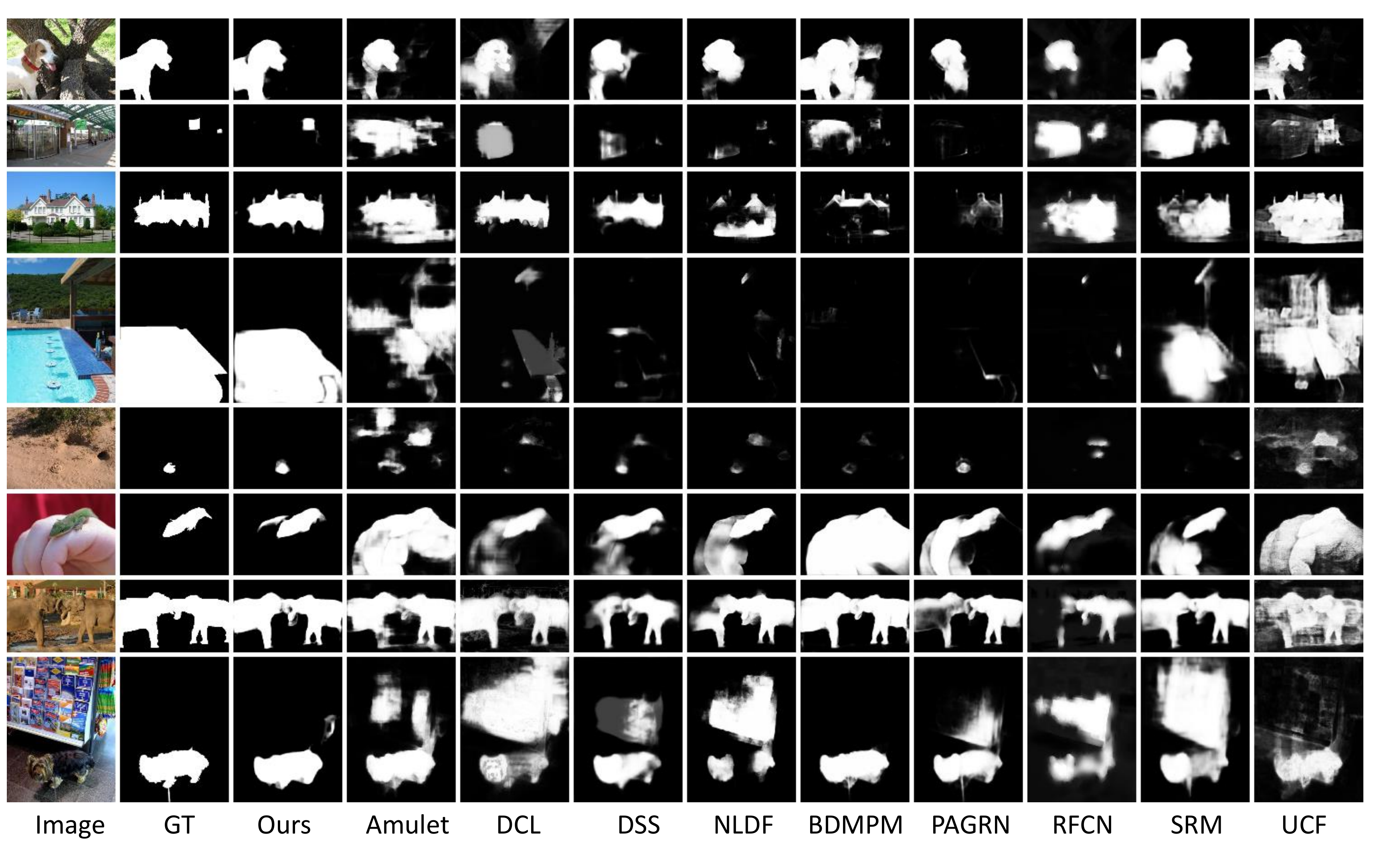}
\caption{ Visual comparisons of the proposed method and the state-of-the-art algorithms.}
\label{fig:Visual_Comparison}
\end{figure*}

\section{Experiments}
\subsection{Datasets and Evaluation Criteria}

The performance evaluation is utilized on five standard benchmark datasets: DUTS-test\cite{duts}, ECSSD\cite{ecssd}, HKU-IS\cite{hkuis}, PASCAL-S\cite{pascal} and DUT-OMRON\cite{omron}.  DUTS\cite{duts} is a large scale dataset, which contains 10553 images for training and 5019 images for testing. ECSSD \cite{ecssd} contains 1,000 images with many semantically meaningful and complex structures in their ground truth segmentation. HKU-IS \cite{hkuis} contains 4447 challenging images with multiple disconnected
salient objects, overlapping the image boundary or low color contrast. PASCAL-S \cite{pascal} contains 850 images, different salient objects are labeled with different saliencies. DUT-OMRON \cite{omron} has 5,168 high quality images. Images of this dataset have one or more salient objects and relatively complex background.

Same as other state-of-the-art salient object detection methods, three popular criteria are used for performance evaluation, i.e. precision and recall curve (denoted PR curve), F-measure, weighted F-measure (denoted $wF_\beta$), and mean absolute error ($MAE$).

The precision and recall are computed by comparing the binary map under different thresholds between predicted saliency map and ground truth, the thresholds are from 0 to 255. $wF_\beta$ is a overall evaluation standard computed by the weighted combination of precision and recall:
\begin{align}
F_\beta = \frac{(1 + \beta^2) \times Precision \times Recall}{\beta^2 \times Precision + Recall}
\end{align}
Where $\beta^2=0.3$ as used in other approaches. Mean absolute error ($MAE$) is computed by:
\begin{align}
MAE = \frac{1}{W \times H} \sum_{x=1}^W \sum_{y=1}^H | P(x,y)-Y(x,y)|
\end{align}
where $Y$ is the ground truth($GT$), and $P$ is the saliency map of network output.

\subsection{Implementation Details}
We use VGG-16 pre-trained on Imagenet\cite{imagenet} as basic model. The DUTS-train dataset is used to train our model, which contains 10553 images. As suggested in \cite{dhs}, we don't use the validation set and train the model until training loss converges. To make the model robust, we adopt some data augmentation techniques: random rotating, random cropping, random brightness, saturation and contrast  changing, and random horizontal flipping.

\begin{center}
\begin{table*}[t]
\tabcolsep=2pt
\centering
\caption{The $wF_\beta$ and $MAE$ of different salient object detection approaches on all test datasets. The best three results are shown in {\color{red}{red}}, {\color{blue}{blue}}, and {\color{green}{green}}.}
\begin{tabular}{| l | c | c | c | c | c | c | c | c | c | c |}
\hline
\multirow{2}*{Methods} & \multicolumn{2}{| c |}{DUTS-test} & \multicolumn{2}{| c |}{ECSSD} & \multicolumn{2}{| c |}{HKU-IS} & \multicolumn{2}{| c |}{PASCAL-S} & \multicolumn{2}{| c |}{DUT-OMRON} \\ \cline{2-11}
&\quad$wF_\beta$\quad\quad & \quad$MAE$\quad&\quad$wF_\beta$\quad\quad & \quad$MAE$\quad&\quad$wF_\beta$\quad\quad & \quad$MAE$\quad&\quad$wF_\beta$\quad\quad & \quad$MAE$\quad&\quad$wF_\beta$\quad\quad & \quad$MAE$\quad\\ \hline
Ours    & {\color{red}0.8702} & {\color{red}0.0405} & {\color{red}0.9313} & {\color{red}0.0328} & {\color{red}0.9264} & {\color{red}0.0324} & {\color{red}0.8922} & {\color{red}0.0677} & {\color{red}0.8557} & {\color{red}0.0414} \\
BDMPM\cite{bdmpm}   & {\color{green}0.8508} & {\color{blue}0.0484} & {\color{blue}0.9249} & {\color{green}0.0478} & {\color{blue}0.9200} & {\color{green}0.0392} & {\color{green}0.8806} & {\color{blue}0.0788} & {\color{green}0.7740} & {\color{green}0.0635} \\
GRL\cite{grl}     & 0.8341 & {\color{green}0.0509} & {\color{green}0.9230} & {\color{blue}0.0446} & 0.9130 & {\color{blue}0.0377} & {\color{blue}0.8811} & {\color{green}0.0799} & {\color{blue}0.7788} & {\color{blue}0.0632} \\
PAGRN\cite{pagrn}   & {\color{blue}0.8546} & 0.0549 & 0.9237 & 0.0643 & {\color{green}0.9170} & 0.0479 & 0.8690 & 0.0940 & 0.7709 & 0.0709 \\
Amulet\cite{amulet}  & 0.7773 & 0.0841 & 0.9138 & 0.0604 & 0.8968 & 0.0511 & 0.8619 & 0.0980 & 0.7428 & 0.0976 \\
SRM\cite{srm}     & 0.8269 & 0.0583 & 0.9158 & 0.0564 & 0.9054 & 0.0461 & 0.8677 & 0.0859 & 0.7690 & 0.0694 \\
UCF\cite{ucf}     & 0.7723 & 0.1112 & 0.9018 & 0.0704 & 0.8872 & 0.0623 & 0.8492 & 0.1099 & 0.7296 & 0.1203 \\
DCL\cite{dcl}     & 0.7857 & 0.0812 & 0.8959 & 0.0798 & 0.8899 & 0.0639 & 0.8457 & 0.1115 & 0.7567 & 0.0863 \\
DHS\cite{dhs}     & 0.8114 & 0.0654 & 0.9046 & 0.0622 & 0.8901 & 0.0532 & 0.8456 & 0.0960 & - & - \\
DSS\cite{dss}     & 0.8135 & 0.0646 & 0.8959 & 0.0647 & 0.9011 & 0.0476 & 0.8506 & 0.0998 & 0.7603 & 0.0751 \\
ELD\cite{eld}     & 0.7372 & 0.0924 & 0.8674 & 0.0811 & 0.8409 & 0.0734 & 0.7882 & 0.1228 & 0.7195 & 0.0909 \\
NLDF\cite{nldf}    & 0.8125 & 0.0648 & 0.9032 & 0.0654 & 0.9015 & 0.0481 & 0.8518 & 0.1004 & 0.7532 & 0.0796 \\
RFCN\cite{rfcn}    & 0.7826 & 0.0893 & 0.8969 & 0.0972 & 0.8869 & 0.0806 & 0.8554 & 0.1159 & 0.7381 & 0.0945 \\\hline
\end{tabular}
\label{tab:all_comparison}
\end{table*}

\end{center}

\begin{figure*}[t]
\centering
\includegraphics[width=1.0\linewidth]{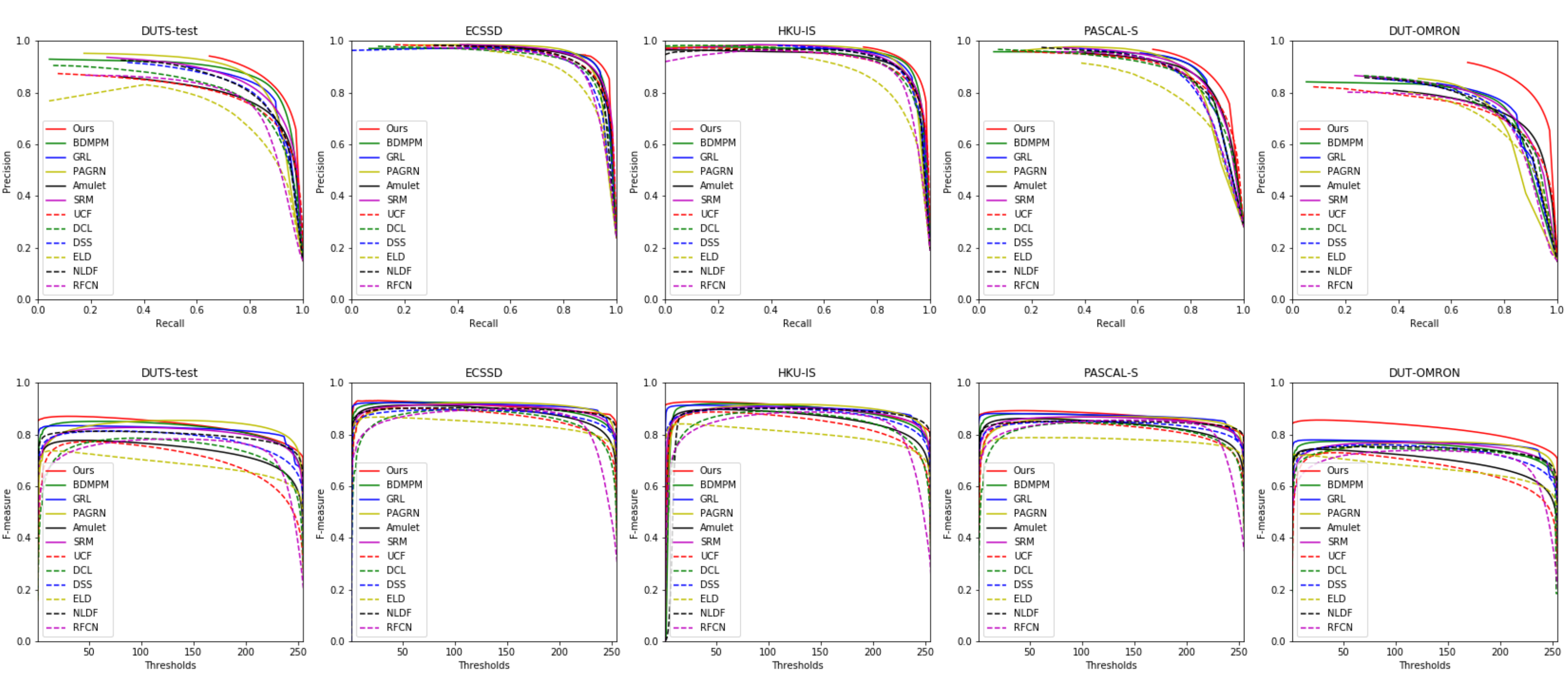}
\caption{ Quantitative comparisons of the proposed approach and eleven state-of-the-art CNN based salient object detection approaches on five datasets. The first and second rows are the PR curves and F-measure curves of different methods respectively.}
\label{fig:Quantitative_Comparison}
\end{figure*}

When training, we set $\alpha$ = 1.0 at beginning to generate rough saliency map. In this period, our model is trained using SGD\cite{sgd} with an initial learning rate 1e-2, the image size is 256$\times$256 , the batch size is 22. Then we adjust different  $\alpha$ to refine the boundaries of saliency map,and find $\alpha$ = 0.7 is the optimal setting in experiment Tab.\ref {tab:bili}. In this period, the image size, batch size is same as the previous period, but the initial learning rate is 1e-3. The code will be found at \url{https://github.com/CaitinZhao/cvpr2019_Pyramid-Feature-Attention-Network-for-Saliency-detection}.

\subsection{Comparison with State-of-the-arts}
The performance of the proposed method is compared with eleven state-of-the-art salient object detection approaches on five test datasets, including BDMPM \cite{bdmpm}, GRL \cite{grl}, PAGRN \cite{pagrn}, Amulet \cite{amulet}, SRM \cite{srm}, UCF \cite{ucf}, DCL \cite{dcl}, DHS \cite{dhs}, ELD \cite{eld}, NLDF \cite{nldf} and RFCN \cite{rfcn}. For fair comparisons, we use the implementations with recommended parameters and the saliency maps provided by the authors.

\subsubsection{Visual Comparison}
Fig.\ref{fig:Visual_Comparison} provides a visual comparison of our method and other state-of-the-arts. From Fig.\ref{fig:Visual_Comparison}, our method gets the best detection results which are much close to the ground truth in various challenging scenarios. To be specific, (1) the proposed method not only highlights the correct salient object regions clearly, but also well suppresses the saliencies of background regions, so as to produce the detection results with higher contrast between salient objects and background than other approaches. (2) With the help of the edge preservation loss, the proposed method is able to generate the salient maps with clear boundaries and consistent saliencies. (3) The saliency maps are much better than other works when salient objects are similar to background (Fig.\ref{fig:Visual_Comparison} the 2,5,7 rows) and the salient objects have special semantic information(Fig.\ref{fig:Visual_Comparison} the 1,3,4,6,8 rows).

\subsubsection{Quantitative Comparison}
Fig.\ref{fig:Quantitative_Comparison} and Tab.\ref{tab:all_comparison} provides the quantitative evaluation results of the proposed method and eleven state-of-the-art salient object detection approaches on five test datasets in terms of PR curve,  F-measure curve, $wF_\beta$ and $MAE$ criteria. As seen from Tab.\ref{tab:all_comparison}, our method gets the best result on five test datasets in terms of $wF_\beta$ and $MAE$, which demonstrate the efficiency of the proposed method. From Fig.\ref{fig:Quantitative_Comparison}, the PR curve and F-measure curve of our method are significantly higher than other methods, which means our method is more robust than other approaches even on challenging datasets. To be specific, our method gets larger improvement compared with the best existing approach on DUT-OMRON dataset. DUT-OMRON dataset is a difficult and challenging saliency detection dataset, in which there are many complex natural scenes images and the color of salient objects is similar to the background. The proposed method can effectively find correct salient objects with powerful feature extraction capability and apt attention mechanisms, which make the network focus on salient objects.

\subsection{The Effectiveness of edge preservation loss}
\begin{figure}[t]
\centering
\includegraphics[width=1.0\linewidth]{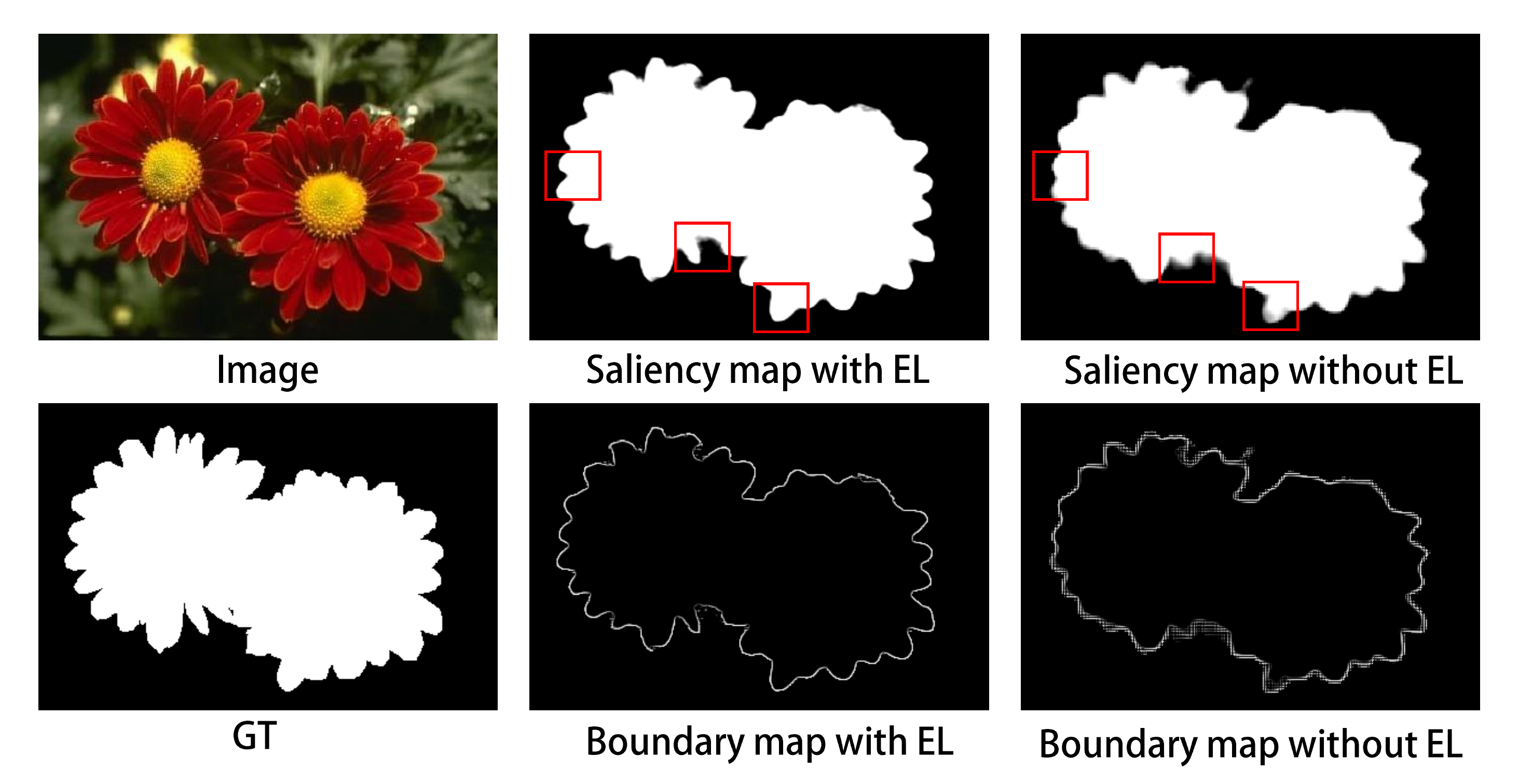}
\caption{Visual comparison of saliency detection results with and without the edge preservation loss.}
\label{fig:edg}
\end{figure}

In Sec.\ref{sec:Loss function} we propose an effective edge preservation loss to guide network to learn more detailed information in boundary localization. Fig.\ref{fig:edg} shows the saliency maps generated from our method and boundary maps calculated by Eq.\ref{abstanh} with edge preservation loss or not. These results illustrate that the edge preservation loss directly enhances the generality and make our method with fine details. In addition, we found that the edge preservation loss with different $\alpha$ have different effects on the final results. From Tab.\ref{tab:bili}, when $\alpha$ is 0.7 gets the best result.

\begin{table}[t]
\begin{center}
\begin{tabular}{|l c c c c c|}
\hline
$\alpha$ & 1. & 0.9 & 0.8 & 0.7 & 0.6 \\\hline
$wF_\beta$ & 0.8528 & 0.8576 & 0.8602 & {\color{red}0.8702} &0.8619 \\
$MAE$ &  0.0432 & 0.0427 & {\color{red}0.0393} & 0.0405 & 0.0428\\
\hline 
\end{tabular}
\end{center}
\caption{The effectiveness of edge preservation loss. The score of $wF_\beta$ and $MAE$ in our method when $\alpha$ is given different values.  The best result is shown in {\color{red}{red}}. The test dataset is DUTS-test. }
\label{tab:bili}
\end{table}

\subsection{Ablation Study}

\begin{table}[h]
\begin{center}
\begin{tabular}{| c c c c c c | c |}
\hline
HL & CPFE & CA & LL & SA & EL & $MAE$  \\ \hline\hline
$\checkmark$&     &    &    &    &    &    0.1003   \\
$\checkmark$&$\checkmark$&    &    &    &    & 0.0815 \\
$\checkmark$&$\checkmark$& $\checkmark$ &    &    &    &  0.0629 \\
$\checkmark$&     &    &$\checkmark$&    &    &  0.0836  \\
$\checkmark$&     &    &$\checkmark$&    &$\checkmark$&  0.0800     \\
$\checkmark$&$\checkmark$&$\checkmark$&$\checkmark$&    &    &  0.0528    \\
$\checkmark$&$\checkmark$&$\checkmark$&$\checkmark$&$\checkmark$&    &  0.0432     \\
$\checkmark$&$\checkmark$&$\checkmark$&$\checkmark$&$\checkmark$&$\checkmark$&  \textbf{0.0405}     \\

\hline
\end{tabular}
\end{center}
\caption{Ablation Study using different components combinations. HL means use High-Level features, CPFE means use Context-aware pyramid Feature Extraction after high-level features, CA means use Channel-wise Attention after high-level features, LL means use Low-Level features, SA means use Spacial Attention after low-level features and EL means use Edge preservation Loss.}
\label{tab:ablation}
\end{table}

To investigate the importance of different modules in our method, we conduct the ablation study. From Tab.\ref{tab:ablation}, that the proposed model contains all components (i.e. context-aware pyramid feature extraction(CPCE), channel-wise attention(CA), spacial attention(SA) and edge preservation loss(EL)) achieves the best performance, which demonstrates that all components are necessary for the proposed method to get the best salient object detection result.

We adopt the model only use high-level features as basic model, and the base $MAE$ is 0.1003. First, we add CPFE to basic model and get decline in $MAE$, furthermore we add CA and get decline of $37\%$ in MAE compared with basic model. Then we add low-level features to high-level features and prove the effectiveness of Integrating multi-scale features. On this basis, we add SA to low-level features and get decline of $57\%$ in MAE compared with basic model. Finally, we add EL in the model and get the best result which get decline of $60\%$ in MAE compared with basic model.

\section{Conclusions}
In this paper, we propose a novel salient object detection method named Pyramid Feature Attention network. In consideration of the different characteristics of different level features, for high-level features we design a context-aware pyramid feature extraction module contains different atrous convolutions at multi scales and a channel-wise attention module to capture semantic high-level features; For low-level features, we employ a spatial attention module to suppress the noises in background and focus on salient objects. Besides, we propose a novel edge preservation loss to guide network to learn more detailed information in boundary localization.  In a word, the proposed method is expert in locating correct salient objects with powerful feature extraction capability and apt attention mechanisms, which make the network robost and effective in saliency detection. Experimental results on five datasets demonstrate that our proposed approach outperforms state-of-the-art methods under different evaluation metrics.

\textbf{Acknowledgments:} This work was supported in part by the Natural Science Foundation of China under Grant 61672194, by the National Key R$\&$D Program of China under Grant 2018YFC0832304, by the Distinguished Youth Science Foundation of Heilongjiang Province of China under Grant JC2018021, and by the State Key Laboratory of Robotics and System (HIT) under Grant SKLRS-2019-KF-14.

{\small
\bibliographystyle{ieee}
\bibliography{egbib}
}

\end{document}